\def\year{2020}\relax
\documentclass[letterpaper]{article} 
\usepackage{aaai20}  
\usepackage{times}  
\usepackage{helvet} 
\usepackage{courier}  
\usepackage[hyphens]{url}  
\usepackage{graphicx} 
\usepackage{graphicx}  
\usepackage{tabularx}
\usepackage{booktabs}
\usepackage{color}
\usepackage{colortbl}
\usepackage{amsmath,amsfonts,amssymb}
\usepackage{subfig}
\usepackage{multirow}
\usepackage[usestackEOL]{stackengine}

\definecolor{mygray}{gray}{.9}

\newcommand{\colordot}[1]{{\textcolor[rgb]{#1}{\LARGE $\bullet$}}}
\newcommand{\para }[1]{\medskip \noindent {\bf #1}}
\newcommand{\tabincell}[2]{\begin{tabular}{@{}#1@{}}#2\end{tabular}}  

\def\Enc{\mathrm{Enc}}
\def\Dec{\mathrm{Dec}}
\def\DecAdv{\mathrm{Dec}^{a}}

\def\INV{{MI}}

\newenvironment{tight_itemize}{
\begin{itemize}
  \setlength{\topsep}{0pt}
  \setlength{\itemsep}{0pt}
  \setlength{\parskip}{0pt}
  \setlength{\parsep}{0pt}
}{\end{itemize}}

\title{Adversarial Learning of Privacy-Preserving and Task-Oriented Representations}
\author{ \Large \textbf{Taihong Xiao\textsuperscript{\rm 1}, Yi-Hsuan Tsai\textsuperscript{\rm 2}, Kihyuk Sohn\textsuperscript{\rm 2~\thanks{Now at Google.}}, Manmohan Chandraker\textsuperscript{\rm 2,3}, Ming-Hsuan Yang\textsuperscript{\rm 1}}\\
$^1$University of California, Merced\\
$^2$NEC Laboratories America\\
$^3$University of California, San Diego\\
}
\begin{document}

\maketitle

\begin{abstract}

Data privacy has emerged as an important issue as data-driven deep learning has been an essential component of modern machine learning systems. For instance, there could be a potential privacy risk of machine learning systems via the model inversion attack, whose goal is to reconstruct the input data from the latent representation of deep networks. Our work aims at learning a privacy-preserving and task-oriented representation to defend against such model inversion attacks. Specifically, we propose an adversarial reconstruction learning framework that prevents the latent representations decoded into original input data. By simulating the expected behavior of adversary, our framework is realized by minimizing the negative pixel reconstruction loss or the negative feature reconstruction (i.e., perceptual distance) loss. We validate the proposed method on face attribute prediction, showing that our method allows protecting visual privacy with a small decrease in utility performance. In addition, we show the utility-privacy trade-off with different choices of hyperparameter for negative perceptual distance loss at training, allowing service providers to determine the right level of privacy-protection with a certain utility performance. Moreover, we provide an extensive study with different selections of features, tasks, and the data to further analyze their influence on privacy protection.
\end{abstract}

\section{Introduction}


As machine learning (ML) algorithms powered by deep neural networks and large data have demonstrated an impressive performance in many areas across natural language~\cite{bahdanau2014neural,wu2016google}, speech~\cite{oord2016wavenet} and computer vision~\cite{krizhevsky2012imagenet,he2016deep}, there have been increased interests in ML-as-a-service cloud services.
These systems demand frequent data transmissions between service providers and their customers to train ML models, or users to evaluate their data. 
For example, customers who want to develop face recognition system may share the set of images containing people of interest with cloud service providers (CSPs). Facial expression based recommendation system may ask users to upload photos.
Unfortunately, these processes are exposed to serious privacy risks. 
Data containing the confidential information shared from the customers can be misused by the CSPs or acquired by the adversary. 
The damage is critical if the raw data is shared with no encryption strategy.


An alternative solution to protect privacy is to encode data using deep features. 
While these approaches are generally more secure than raw data, recent advances in model inversion (\INV) techniques~\cite{mahendran2015understanding,dosovitskiy2016inverting,zhang2016augmenting,dosovitskiy2016generating} call the security of deep features into question. 
For example, \cite{dosovitskiy2016generating} shows that adding deep image prior (e.g., perceptual similarity) allows inversion from low-, mid-, as well as high-level features.

\begin{figure}[t]
    \centering
    \includegraphics[width=.99\linewidth]{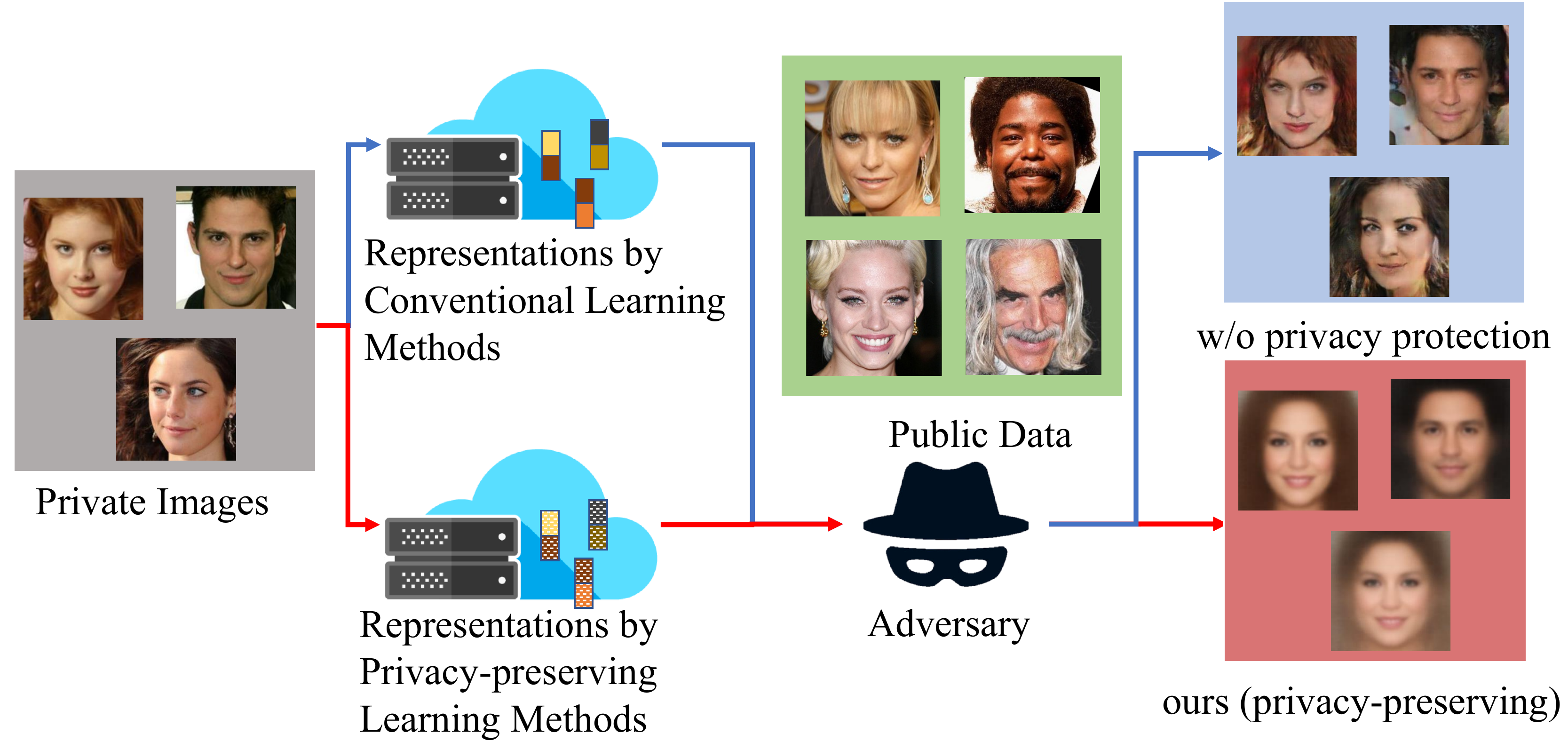}
    \caption{
    Representations of conventional deep learning algorithms are vulnerable to adversary's \emph{model inversion attack}~\cite{fredrikson2015model}, which raise serious issues on data privacy. Our method learns task-oriented features while preventing private information being decoded into an input space by simulating the adversary's behavior at training phase via negative reconstruction loss. 
    %
    }
	\label{fig:teaser}
	\vspace{-2mm}
\end{figure}


In this work, we study how to learn a privacy-preserving and task-oriented deep representation. 
We focus on the defense against \emph{a black-box model inversion attack}, where the adversary is allowed to make unlimited inferences\footnote{While our assumption is the most favorable scenario to adversary, in practice, the CSPs can limit the number of inferences. {\INV} attack with limited inference is beyond the scope of this work.} of their own data\footnote{Although we assume the adversary has no direct access to the private data used for model training, adversary's own data is assumed to be representative of them.} to recover input from acquired features of private customer and user data. 
As an adversary, we consider an {\INV} attack based on the neural decoder that is trained to reconstruct the data using data-feature pairs. Perceptual~\cite{dosovitskiy2016generating} and GAN~\cite{goodfellow2014generative} losses are employed to improve the generation quality. 
Finally, we present our \emph{adversarial data reconstruction learning} that involves an alternate training of encoder and decoder. 
While the decoder, simulating the adversary's attack, is trained to reconstruct input from the feature, the encoder is trained to \emph{maximize} the reconstruction error to prevent decoder from inverting features while minimizing the task loss.

We explain the vulnerability of deep networks against the {\INV} attacks in the context of facial attribute analysis with extensive experimental results. 
We show that it is difficult to invert the adversarially learned features and thus the proposed method successfully defends against a few strong inversion attacks. 
In this work, we perform extensive experiments by inverting from different CNN layers, with different data for adversary, with different utility tasks, with different weight of loss term, to study their influences on data privacy. 
Furthermore, we show the effectiveness of our method against feature-level privacy attacks by demonstrating the improved invariance on the face identity, even when the model is trained with no identity supervision.

The contributions of this work are summarized as follows:
\begin{tight_itemize}
    \item We propose an adversarial data reconstruction learning to defend against black-box model inversion attacks, along with a few strong attack methods based on neural decoder.
    \item We demonstrate the vulnerability of standard deep features and the effectiveness of the features learned with our method in preventing data reconstruction.
    \item We show the utility-privacy trade-off with different choice of hyperparameter for negative perceptual distance loss.
    \item We perform extensive study of the impacts on the privacy protection with different layers of features, tasks, and data for decoder training.
\end{tight_itemize}

\section{Related Work}

\subsection{Data Privacy Protection}
To protect data privacy, numerous approaches have been proposed based on information theory~\cite{liang2009information}, statistics~\cite{du2012privacy,smith2011privacy}, and learnability~\cite{kasiviswanathan2011can}. Furthermore, syntactic anonymization methods including $k$-anonymity~\cite{sweeney2002k}, $l$-diversity~\cite{machanavajjhala2006ell} and $t$-closeness~\cite{li2007t} are developed. Nevertheless, these approaches protect sensitive attributes in a static database but do not scale well to high-dimensional image data. 
On the other hand, the concept of differential privacy~\cite{dwork2004privacy,dwork2006calibrating} has been introduced to provide formal privacy guarantee. It prevents an adversary from gaining additional knowledge by including or excluding an individual subject, but the information leaked from the released data itself is not discussed in these works. 
\subsection{Visual Privacy Protection}
Typical privacy-preserving visual recognition methods aim to transform the image data such that identity cannot be visually determined, based on image operations such as Gaussian blur~\cite{oh2016faceless}, mean shift filtering~\cite{winkler2014trusteye}, down-scaling, identity obfuscation~\cite{oh2016faceless}, and adversarial image perturbation~\cite{Oh_2017_ICCV}. Although effective in protecting privacy, these methods have negative impact on utility. 
To overcome this limitation, numerous algorithms have been proposed to complete the utility task based on transformed data. \cite{wang2016studying} design a method to improve low-resolution recognition performance via feature enhancement and domain adaptation. In \cite{ryoo2017privacy}, it is demonstrated that reliable action recognition may be achieved at low resolutions by learning appropriate down-sampling transformations.
Furthermore, trade-offs between resolution and action recognition accuracy are discussed in \cite{dai2015towards}. Furthermore, \cite{Oh_2017_ICCV} propose an adversarial method to learn the image perturbation so as to fool the face identity classifier, but the adversarial perturbed images are visually exposing the identity privacy. Different from these methods, our method learns image features so as to protect the privacy, which could also maintain the utility performance to some extent.

\begin{figure*}[t]
    \centering
    \subfloat[Update {\color{blue}$\Dec$} and {\color{blue}$D$} using $X\,{\in}\,\mathcal{X}_{2}$ while fixing $\Enc$ and $f$.]{\includegraphics[width=0.45\textwidth]{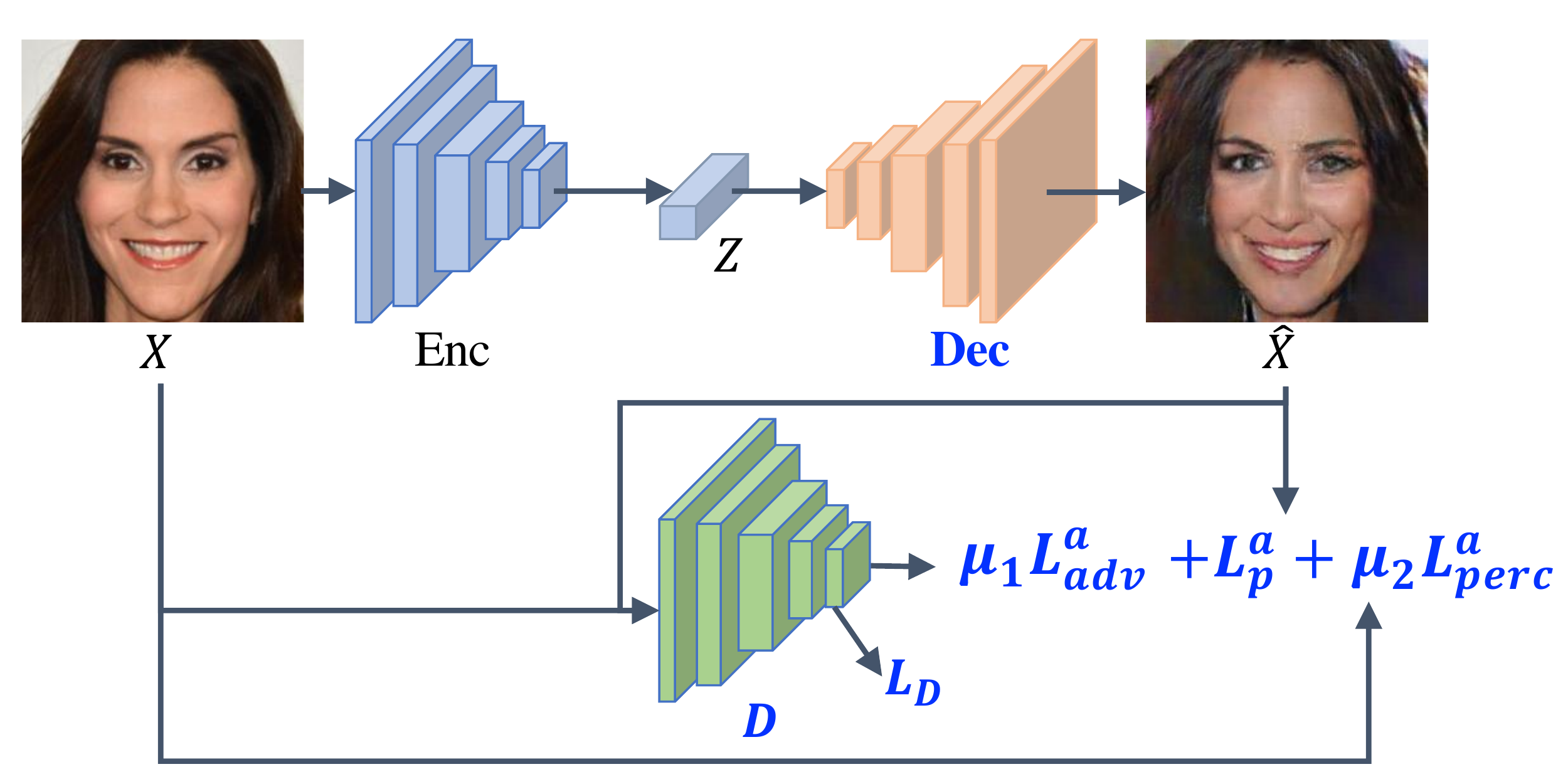}\label{fig:framework_dec}}
    \hfil
    \subfloat[Update {\color{blue}$\Enc$} and {\color{blue}$f$} using $X\,{\in}\,\mathcal{X}_{1}$ while fixing $\Dec$.]{\includegraphics[width=0.45\textwidth]{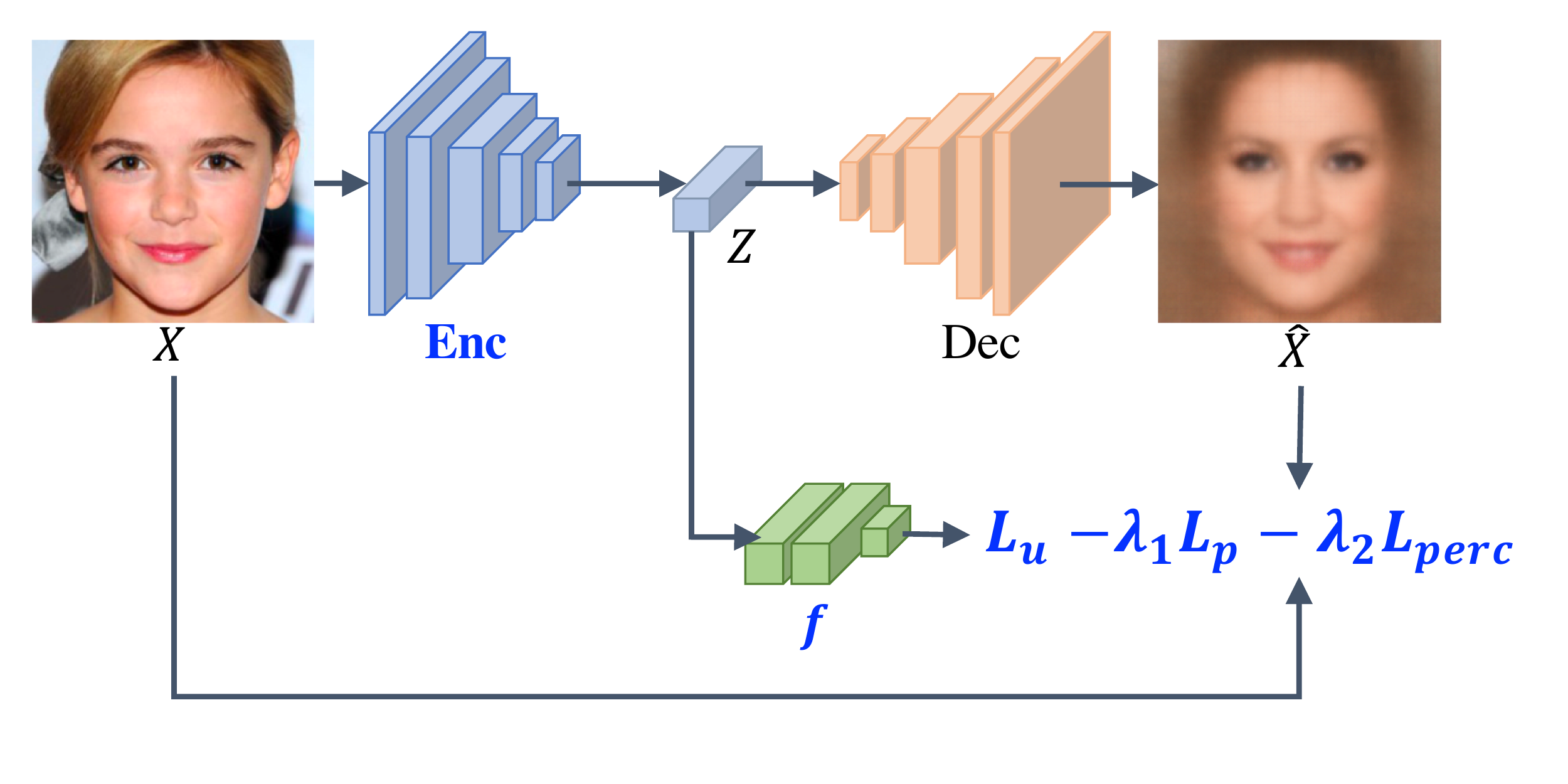}\label{fig:framework_enc}}
    \caption{An overview of our privacy-preserving representation learning method with adversarial data reconstruction.
    As in (a), decoder (Dec) is trained to reconstruct an input $X$ from latent encoding $Z=\Enc(X)$ on public data $\mathcal{X}_{2}$.
    Then, as in (b), we update Enc on private data $\mathcal{X}_{1}$ to generate $Z$ that fools Dec, i.e., prevent reconstructing an input $X$ by Dec, while achieving utility prediction performance via the classifier $f$.}
	\label{fig:framework}
	\vspace{-2mm}
\end{figure*}

\subsection{Feature and Model Inversion}
There has been a growing interest in stealing the functionality from the black-box classifier~\cite{Orekondy2019cvpr} or inverting CNNs to better understand what is learned in neural networks. \cite{mahendran2015understanding} introduce optimization-based inversion technique that allows low- and mid-level CNN features to be inverted. 
\cite{dosovitskiy2016inverting,zhang2016augmenting} suggest to invert via up-convolution neural network and demonstrate improved network inversion from mid-level representations.
Nevertheless, this method is less effective in inverting high-level features.
More importantly, recent studies~\cite{dosovitskiy2016generating} show that adding perceptual image prior allows inversion from high-level representations possible, exposing a potential privacy issue by inverting deep representations. 
However, our method prevents image representations from being reconstructed to original images via adversarial learning.

\subsection{Trade-off between Privacy and Utility}
Recently, several methods protect the utility performance when the information required for the utility task is unrelated to the private information. 
For example, \cite{Wu_2018_ECCV,Ren_2018_ECCV} disguise original face images (privacy) without sacrificing the performance of action recognition (utility), since the information being useful to action recognition is independent of the face information. 
However, there still remains a question of how to better protect the privacy when the information for the utility task is related to the private information. 
%
In \cite{Pittaluga2018privacy}, an encoding function is learned via adversarial learning to prevent the encoded features from making predictions about the specific attributes. Differently, our method does not require additional annotations for the private attributes and instead resolves this issue via an adversarial reconstruction framework.

\subsubsection{Differences between Our method and Existing methods.}
In contrast to prior work, our approach can 1) scale up to high-dimensional image data compared with those methods based on $k$-anonymity, $\ell$-diversity abd $t$-closeness that are often applied to field-structured data (e.g., tabular data); 2) learn image representations as the alternative of raw images while soem approaches focus on removing private information in raw images; 3) require no definition of private attribute or entailment of additional annotations.

\section{Proposed Algorithm}

Before introducing our privacy-preserving feature learning method, we discuss the privacy attack methods of the adversary. 
We then describe the proposed privacy-preserving and task-oriented representation learning.


\subsection{Adversary: Learn to Invert}
\label{sec:update_dec}
To design a proper defense mechanism, we first need to understand an adversary's attack method. Specifically, we focus on the \emph{model inversion (\INV) attack}~\cite{fredrikson2015model}, where the goal of the adversary is to invert features back to an input. We further assume a \emph{black-box} attack, where the adversary has unlimited access to the model’s inference ($\Enc$), but not the model parameters. This is extremely generous setting to the adversary since they can create a large-scale paired database of input $X\,{\in}\,\mathcal{X}_{2}$ and the feature $Z\,{=}\,\Enc(X)$. Here, we use $\mathcal{X}_{2}$ to distinguish the adversary's own dataset from the private training dataset $\mathcal{X}_{1}$ of CSPs or their customers. 

Given a paired dataset $\{(X{\in}\mathcal{X}_{2},Z)\}$, the adversary inverts the feature via a decoder $\DecAdv\,{:}\,\mathcal{Z}\rightarrow\mathcal{X}$, which is trained to reconstruct the input $X$ from the feature $Z$ by minimizing the reconstruction loss\footnote{Optimization-based inversion attack~\cite{mahendran2015understanding} may be considered instead, but it is not feasible in the black-box setting since the adversary has no access to $\Enc$’s model parameters. Even in the white-box scenario, inversion by decoder may be preferred as it is cheaper to compute.}:
\begin{align}
    \mathcal{L}^{a}_{p} = \mathbb{E}_{\{(X{\in}\mathcal{X}_{2},Z)\}}\big[\Vert\hat{X} - X\Vert^{2}\big],\label{eq:update_dec}
\end{align}
where $\hat{X}\,{=}\,\DecAdv(Z)$. Note that the above does not involve backpropagation through an $\Enc$. The inversion quality may be improved with the GAN loss~\cite{goodfellow2014generative}:
\begin{align}
    \mathcal{L}^{a}_{\text{adv}} & = \mathbb{E}_{Z}\big[\log\big(1{-}D(\hat{X})\big)\big],\label{eq:gan_loss}
\end{align}
where $D$ is the discriminator, telling the generated images from the real ones by maximizing the following loss:
\begin{align}
    \mathcal{L}^{a}_{D} & = \mathbb{E}_{\{(X{\in}\mathcal{X}_{2}, Z)\}}\big[\log\big(1{-}D(X)\big)+\log\big(D(\hat{X})\big)\big].\label{eq:gan_loss_discriminator}
\end{align}
We can further improve the inversion quality by minimizing the perceptual distance~\cite{dosovitskiy2016generating}:
\begin{equation}\label{eq:perceptual_similarity}
    \mathcal{L}^{a}_{\text{perc}} = \mathbb{E}_{\{(X{\in}\mathcal{X}_{2},Z)\}} \big[\Vert g(\DecAdv(Z)) - g(X)\Vert^{2}\big],
\end{equation}
where we use the conv1 to conv5 layers of the VGG network~\cite{simonyan2014very} pre-trained on the ImageNet
for $g$. 
The overall training objective of an adversary thus can be written as:
\begin{equation}\label{eq:adversary_objective}
\left\{
\begin{array}{ll}
    {\min\limits_{\DecAdv}} \; & \mathcal{L}^{a}_{p} + \mu_{1}\mathcal{L}^{a}_{\text{adv}} + \mu_{2}\mathcal{L}^{a}_{\text{perc}},\\
    {\max\limits_{D}}\; & \mathcal{L}^{a}_{D}.
\end{array}
\right.
\end{equation}

\begin{table*}[!t]
\def\arraystretch{1.1}
\setlength\tabcolsep{5pt}
  \centering
  \caption{Results on facial attribute prediction. $\lambda_{2}$, and $\mu_{2}$ are fixed to 0. We report the MCC averaged over 40 attributes as a utility metric and face and feature similarities as privacy metrics. The $\Enc$ is trained using binary cross entropy (BCE) without or with the proposed adversarial reconstruction loss. Rows with $^\dagger$ (\#5 and \#6) train $\DecAdv$ with an extra data from MS-Celeb-1M, while those with $^\ddagger$ (\#7 and \#8) use an extra data to train both $\Enc$ and $\DecAdv$. Different $\mu_{1}$'s in~\eqref{eq:perceptual_similarity} are deployed for {\INV} attack. \#9 and \#10 consider {\it Smiling} attribute prediction for utility task and MCC is evaluated only for Smiling attribute.}
  \footnotesize
    \begin{tabular}{l|c|c|c|cc|cc}
    \toprule
    ID & $\Enc$ & $\DecAdv$ & \multicolumn{1}{c|}{Mean MCC $\uparrow$} & \multicolumn{1}{c}{Face Sim. $\downarrow$} & \multicolumn{1}{c|}{Feature Sim. $\downarrow$} & \multicolumn{1}{c}{SSIM} & \multicolumn{1}{c}{PSNR} \\
    \midrule
    1 & $\lambda_{1}=0$ & $\mu_{1}=0$ & 0.641 & 0.551 & 0.835 & 0.231 & 13.738   \\
    \rowcolor{mygray}
    2 & $\lambda_{1}>0$ & $\mu_{1}=0$ & 0.612 & 0.515 & 0.574 & 0.221 & 13.423 \\
    3 & $\lambda_{1}=0$ & $\mu_{1}>0$ & 0.641 & 0.585 & 0.835 & 0.240  & 14.065 \\
    \rowcolor{mygray}
    4 & $\lambda_{1}>0$ & $\mu_{1}>0$ & 0.612 & 0.513 & 0.574  & 0.277  & 13.803   \\
    \midrule
    \multicolumn{8}{c}{With more data for training $\DecAdv$ (ID \#5 and \#6) and both $\Enc$ and $\DecAdv$ (ID \#7 and \#8)} \\
    \midrule
    5 & $\lambda_{1}=0^{\dagger}$ & $\mu_{1}=0$ & 0.641 & 0.594 & 0.864 & 0.250  & 14.132  \\
    \rowcolor{mygray}
    6 & $\lambda_{1}>0^{\dagger}$ & $\mu_{1}=0$ & 0.612 & 0.541 & 0.633 & 0.222  & 13.703  \\
    7 & $\lambda_{1}=0^{\ddagger}$ & $\mu_{1}=0$ & 0.651 & 0.579 & 0.834  & 0.263  & 14.432  \\
    \rowcolor{mygray}
    8 & $\lambda_{1}>0^{\ddagger}$ & $\mu_{1}=0$ & 0.630 & 0.550 & 0.591 & 0.231  & 13.334  \\
    \midrule
    \multicolumn{8}{c}{Single (Smiling) attribute prediction. MCC for Smiling attribute is reported in the parenthesis.} \\
    \midrule
    9 & $\lambda_{1}=0$ & $\mu_{1}>0$ & 0.001 (0.851) & 0.460 & 0.494  & 0.204  & 13.214  \\
    \rowcolor{mygray}
    10 & $\lambda_{1}>0$ & $\mu_{1}>0$ & 0.044 (0.862) & 0.424 & 0.489 &  0.189  & 12.958 \\
    \bottomrule
    \end{tabular}
  \label{tab:40_attributes_classification}
\end{table*}

\subsection{Protector: Learn ``NOT'' to Invert}
\label{sec:update_enc_f}

Realizing the attack types, we are now ready to present the training objective of privacy-preserving and task-oriented representation. To learn a task-oriented representation, we adopt an MLP classifier $f$ that predicts the utility label $Y$ from $Z$ by minimizing the utility loss:
\begin{equation}\label{eq:utility_function} 
\mathcal{L}_{u} = \mathbb{E}_{\{(X{\in}\mathcal{X}_{1},Y)\}}\big[\mathcal{L}(f(Z), Y)\big],
\end{equation}
where $f(Z)\,{=}\,f(\Enc(X))$, $Y$ is the ground-truth label for utility, and $\mathcal{L}$ is the standard loss (e.g., cross-entropy) for utility. Note that $\mathcal{X}_{1}$ is a private training data, which is not accessible to the adversary.

To learn a privacy-preserving feature against the {\INV} attack, an $\Enc$ needs to output $Z$ that cannot be reconstructed into an input $X$ by \emph{any} decoder. Unfortunately, enumerating all possible decoders is not feasible. Instead, we borrow the idea from adversarial learning~\cite{goodfellow2014generative}. To be more specific, the decoder ($\Dec$) is trained to compete against an $\Enc$ in a way that it learns to decode $Z$ of the current $\Enc$ into $X$ by minimizing the reconstruction loss:
\begin{align}
\mathcal{L}_{p} = \mathbb{E}_{\{(X{\in}\mathcal{X}_{1},Z)\}}\big[\Vert \Dec(Z)\,{-}\,X\Vert^2\big].
\label{eq:negative_reconstruction_loss}
\end{align}
%
In addition, one can improve the quality of reconstruction, resulting in a stronger adversary, using perceptual distance loss as in~\eqref{eq:perceptual_similarity}:
\begin{align}
\mathcal{L}_{\text{perc}} & = \mathbb{E}_{\{(X{\in}\mathcal{X}_{1},Z)\}}\big[\Vert g(\Dec(Z))\,{-}\,g(X)\Vert^2\big].\label{eq:negative_perceptual_loss}
\end{align}
On the other hand, an $\Enc$ aims to fool $\Dec$ by maximizing the reconstruction loss or perceptual distance loss. Finally, the overall training objective of a protector is:
\begin{equation}\label{eq:protector_formula}
    \min\limits_{\Enc, f} \; \mathcal{L}_{u} - \lambda_{1}\mathcal{L}_{p} - \lambda_{2}\mathcal{L}_{\text{perc}},
\end{equation}
while the decoder $\Dec$ is updated by the same loss function as $\DecAdv$ according to \eqref{eq:adversary_objective}.
%
%
%
Figure~\ref{fig:framework} shows the main modules of our method. We adopt alternative update strategy for the proposed learning framework. The $\DecAdv$ and $D$ are updated first on the public data $\mathcal{X}_2$ according to \eqref{eq:adversary_objective} whiling fixing the $\Enc$ and $f$, and $\Enc$ and $f$ are updated on the private data by \eqref{eq:protector_formula} and so forth until convergence.

\section{Experimental Results}

We first introduce two types of privacy attack methods: the black-box MI attack and feature-level attack, and two metrics: face similarity and feature similarity. 
Next, we show the effectiveness of our method against privacy attacks in different scenarios. 


\begin{figure*}[!t]
\centering
\def\arraystretch{0.9}
\def\tabuwidth{0.11\linewidth}
\begin{tabular}{m{\tabuwidth}m{0.015\linewidth}*{5}{m{\tabuwidth}}}
    \multicolumn{1}{c}{} &       & \multicolumn{1}{c}{$\mu_1=0$} & \multicolumn{1}{c}{$\mu_1>0$} &  \multicolumn{1}{c}{$\mu_1=0^{(\dagger)}$} & \multicolumn{1}{c}{$\mu_1=0^{(\ddagger)}$} & \multicolumn{1}{c}{$\mu_{1}>0\textsuperscript{(Smiling)}$} \\
    \multirow{2}[0]{*}{\tabincell{c}{input\\\includegraphics[width=\linewidth]{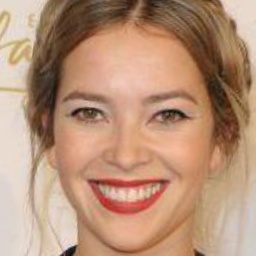}}} & {\centering \rotatebox{90}{$\lambda_{1}=0$}}
    & \includegraphics[width=\linewidth]{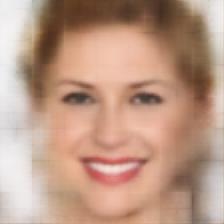}  
    & \includegraphics[width=\linewidth]{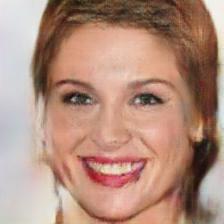}  
    & \includegraphics[width=\linewidth]{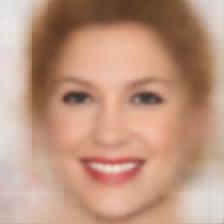}  
    & \includegraphics[width=\linewidth]{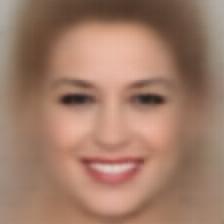}
    & \includegraphics[width=\linewidth]{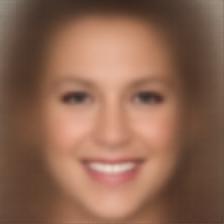} \\
    & \rotatebox{90}{$\lambda_{1}>0$} 
    & \includegraphics[width=\linewidth]{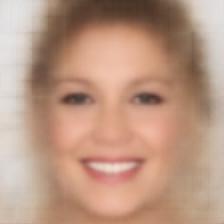}  
    & \includegraphics[width=\linewidth]{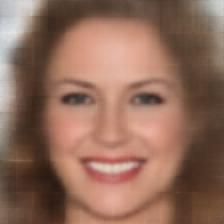}  
    & \includegraphics[width=\linewidth]{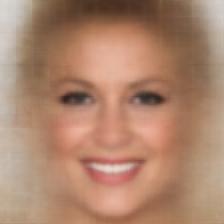}
    & \includegraphics[width=\linewidth]{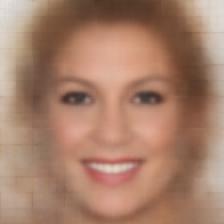} 
    & \includegraphics[width=\linewidth]{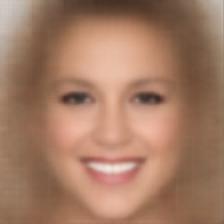} \\
\end{tabular}%
\vspace{-2mm}
\caption{Visualization of reconstructions by $\DecAdv$ under various $\Enc$ and $\DecAdv$ settings. $\lambda_{2}$ and $\mu_{2}$ are fixed to 0.
Examples in the second row ($\lambda_{1} > 0$) are results using our negative reconstruction loss, which shares less similarities to the input than the examples in the first row, where the negative reconstruction loss is not employed. 
}
\label{fig:adversarial_comparison}
\end{figure*}

\begin{figure*}[!t]
\vspace{-1mm}
\def\pwidth{0.18\textwidth}
\centering
\begin{tabular}{c|c|c|c}
\toprule
\multicolumn{2}{c|}{40 Attributes} & \multicolumn{2}{c}{Single ({\it Smiling}) Attribute} \\
\midrule
w/o adv & w/ adv & w/o adv & w/ adv \\

\includegraphics[width=\pwidth]{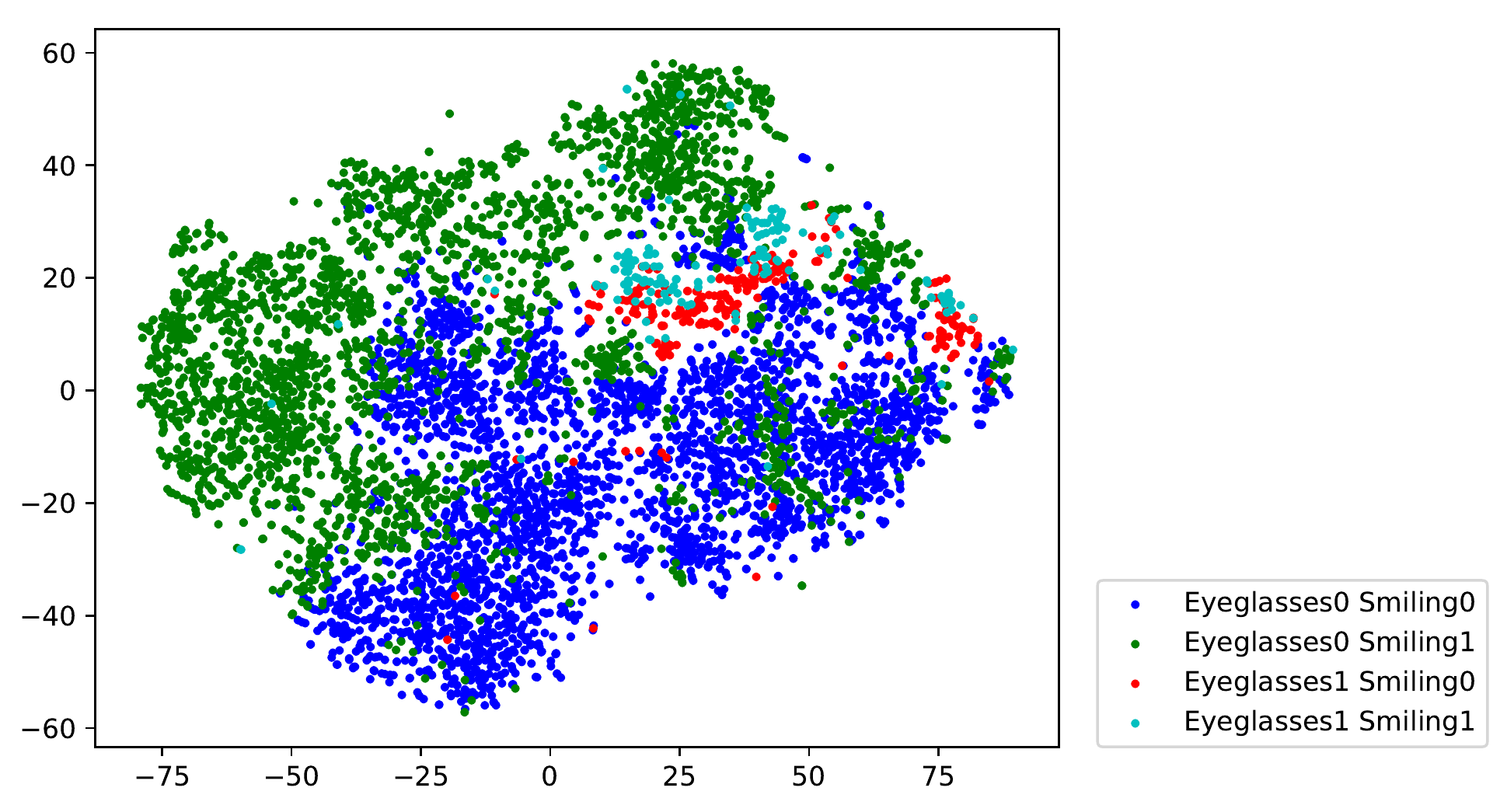}&
\includegraphics[width=\pwidth]{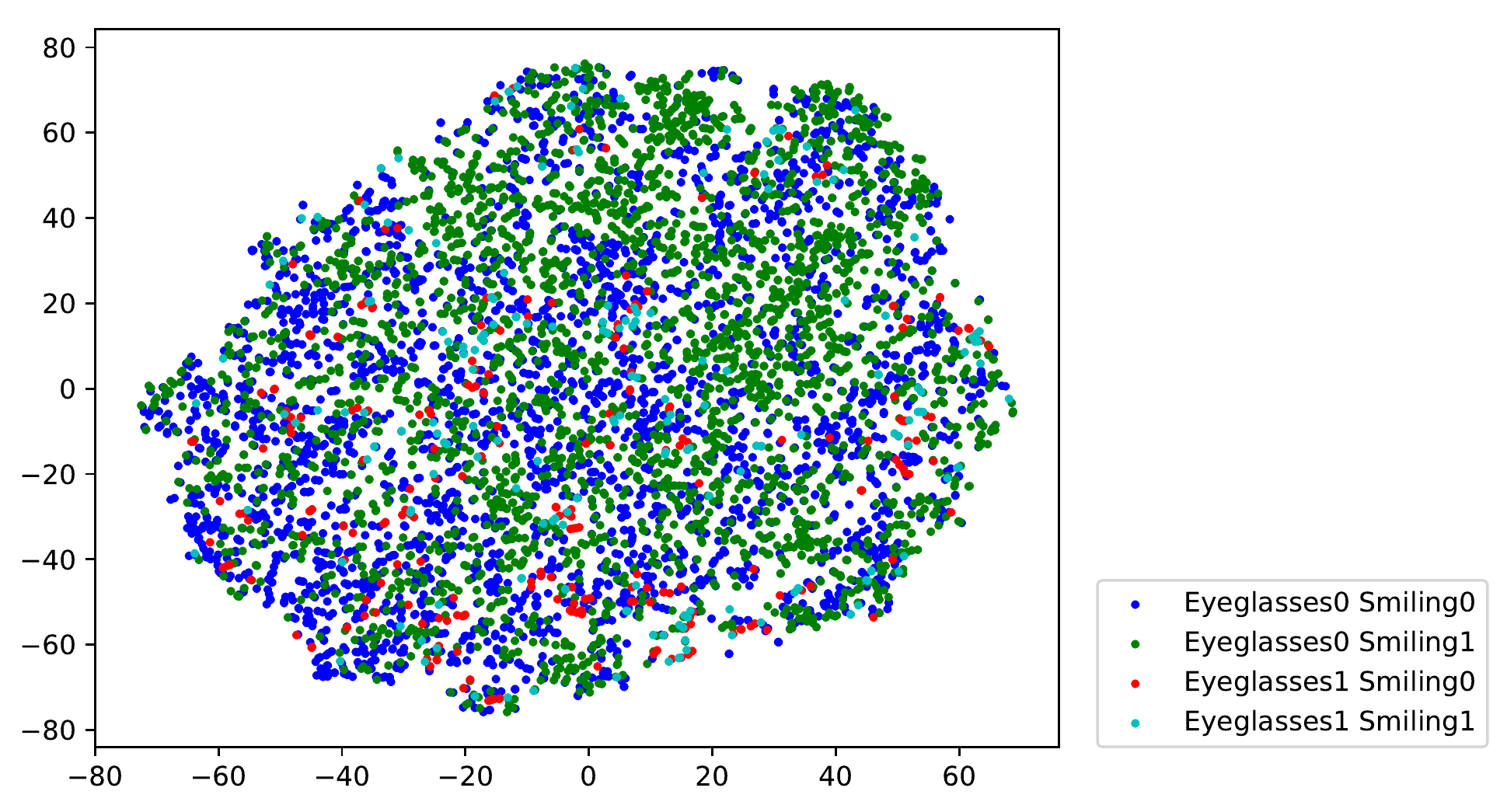}&
\includegraphics[width=\pwidth]{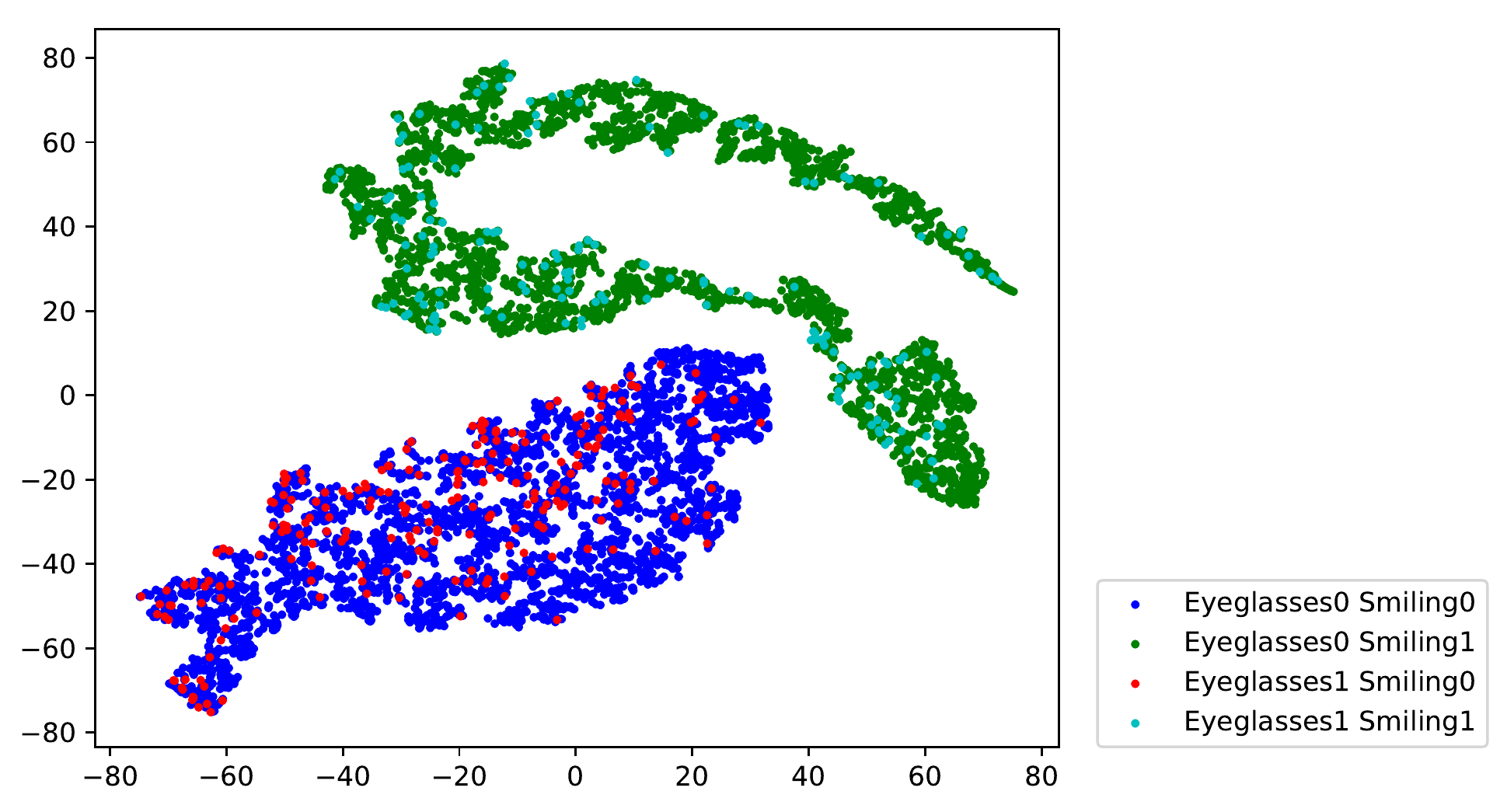}&
\includegraphics[width=\pwidth]{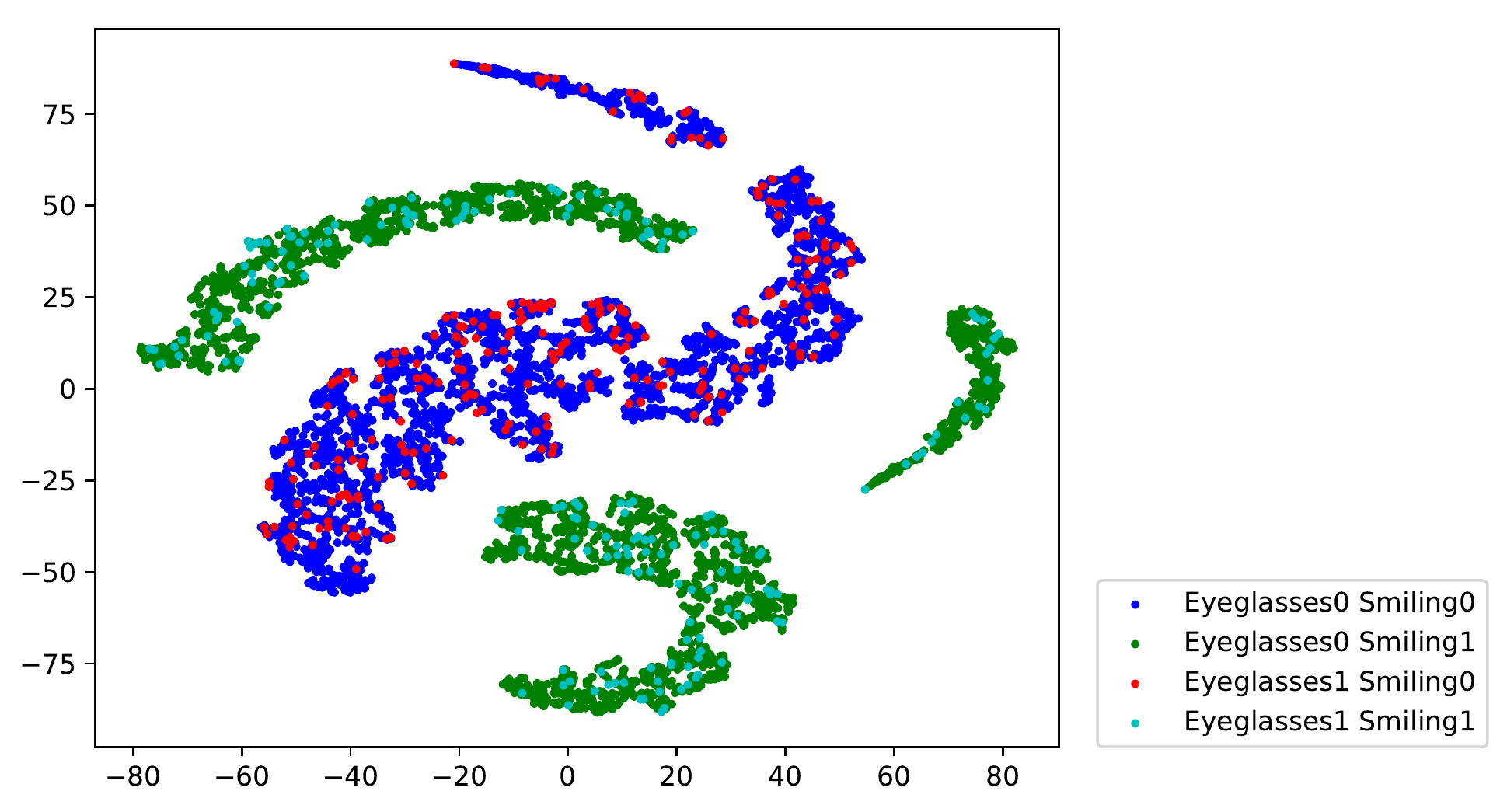}\\
\bottomrule
\end{tabular}

\begin{tabular}{cccc}
\colordot{0,0,1} {\scriptsize Eyeglasses$-$ Smiling$-$} & \colordot{0,1,0} {\scriptsize Eyeglasses$-$ Smiling$+$} & \colordot{1,0,0} {\scriptsize Eyeglasses$+$ Smiling$-$} & \colordot{0,1,1} {\scriptsize Eyeglasses$+$ Smiling$+$}\\
\end{tabular}
\vspace{-2mm}
\caption{Visualization of reconstructed images from latent representations. ``w/o adv'' represents reconstruction from conventional latent representations, whereas ``w/ adv'' means reconstruction from our privacy-preserving representations. All points can be categorized into 4 classes (i.e., each attribute has $+$ or $-$) in different colors as illustrated above. 
}
\label{fig:latent_visualization}
\vspace{-2mm}
\end{figure*}

\subsection{Dataset and Experimental Setup}

We use the widely-used CelebA~\cite{liu2015faceattributes} and MS-Celeb-1M~\cite{guo2016msceleb} datasets for experiments. 
In most experiments, we split the CelebA dataset into three parts, $\mathcal{X}_{1}$ with 160k images, $\mathcal{X}_{2}$ with 40k images, and the test set $\mathcal{T}$ with the rest. 
For ablation studies with different data for adversaries, we provide extra data for $\mathcal{X}_{2}$ from the MS-Celeb-1M dataset. 
%
%
We use the ResNet-50 model~\cite{he2016deep} for feature representation and two fully connected layers for latent classifier $f$.
The $\Dec$ uses up-sampling layers to decode features to pixel images.
We compare the proposed adversarial feature learning method to the baseline where $\Enc$ is trained to minimize the binary cross entropy (BCE) for prediction for $40$ binary facial attributes. 
More details regarding the networks can be found in the supplementary material.

%
%
%

%
%
%

\subsection{Evaluation against Data Privacy Attacks}

\para{Adversary's Attacks.} Given an $\Enc$, we evaluate the robustness of its feature by simulating the adversary's attacks, i.e., the {black-box {\INV} attack} and the {feature-level attack}.

While the {\INV} attack aims at reconstructing the input to recover the holistic information, the feature-level attack aims to recover predefined private attributes from the feature. 
In other words, the adversary aims to learn a mapping function $M\,{:}\,\mathcal{Z}\rightarrow\mathcal{C}$ to reconstruct the feature of private attribute prediction network $C$ (e.g., face verification network):
\begin{align}
    \min_{M} \mathbb{E}_{X\in\mathcal{X}_2}\Big[\Vert M(Z) - C(X)\Vert^2\Big].
    \label{eq:fea_attack}
\end{align}
The privacy information is not well protected if one finds $M$ that successfully minimizes the loss in~\eqref{eq:fea_attack}.


\para{Utility Metric.}
We measure the attribute prediction performance of $f\circ\Enc$ on $\mathcal{T}$.
Due to the imbalanced label distribution of the CelebA dataset, we use the Matthews correlation coefficient (MCC)~\cite{boughorbel2017optimal} as the evaluation metric:
\begin{align}
\mbox{MCC} = \frac{(\mbox{TP}\cdot \mbox{TN} - \mbox{FP}\cdot \mbox{FN})}{\sqrt{(\mbox{TP+FP})(\mbox{TP+FN})(\mbox{TN+FP})(\mbox{TN+FN})}},\nonumber
\end{align}
where TP and FN stand for true positive and false negative, respectively. 
The MCC value falls into a range of $[-1, +1]$, in which $+1$, $0$, and $-1$ indicate the perfect, random, and the worst predictions, respectively.

\para{Privacy Metric.}
For the {\INV} attack, we compare the reconstruction $\hat{X}\,{=}\,\Dec^a(\Enc(X))$ to $X$ by visual inspection and perceptual similarity. 
The face similarity between $X$ and $\hat{X}$ is computed by the cosine similarity of their deep features, e.g., layer 3 of $C$. 
%
Here, we use identity as a private attribute and an OpenFace face verification model~\cite{amos2016openface} for $C$. 
We also report the SSIM and PSNR of the reconstructed images to evaluate the realism.
For the feature-level attack, we report the cosine similarity of features between $M(Z)$ and $C(X)$.
%
%

\subsection{Empirical Results and Performance Analysis}
\para{Baseline.}
In Table~\ref{tab:40_attributes_classification}, we first present main results by fixing $\mu_2$ and $\lambda_2$ equal to 0. The baseline model trained only with the binary cross entropy (ID \#1, \#3) of 40 attributes are compared with our model trained with negative reconstruction loss, i.e., $\lambda_1 = 1$ (ID \#2, \#4). 
We simulate the weak and strong adversaries without ($\mu_{1}\,{=}\,0$) or with ($\mu_{1}\,{>}\,0$) the GAN loss for training $\DecAdv$. 
The proposed model maintains a reasonable task performance with a decrease in face similarity and feature similarity (\#1 vs \#2, \#3 vs \#4 in Table~\ref{tab:40_attributes_classification}). 
Such improved privacy protection is also shown in face or feature similarities and visual inspection (Figure~\ref{fig:adversarial_comparison}).

\begin{figure*}[!t]
\centering
\def\arraystretch{1}
\begin{tabular}{*{8}{m{0.1\linewidth}<{\centering}}}
    input & 0 & 1 & 2 & 3 & 4 & 5 & 10 \\
    \includegraphics[width=\linewidth]{190020}
    & \includegraphics[width=\linewidth]{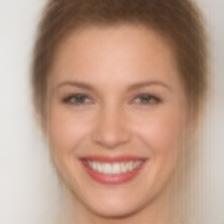} 
    & \includegraphics[width=\linewidth]{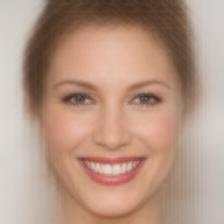}  
    & \includegraphics[width=\linewidth]{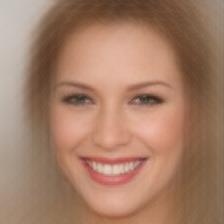}
    & \includegraphics[width=\linewidth]{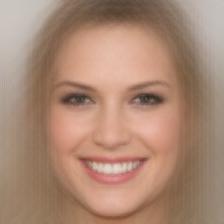}
    & \includegraphics[width=\linewidth]{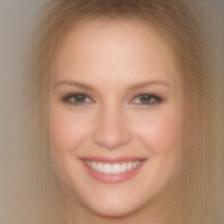}
    & \includegraphics[width=\linewidth]{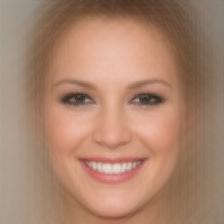}
    & \includegraphics[width=\linewidth]{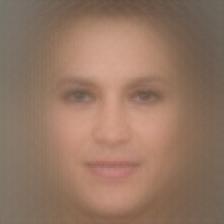} \\
    
    \includegraphics[width=\linewidth]{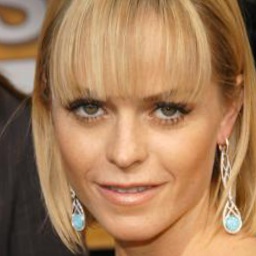}
    & \includegraphics[width=\linewidth]{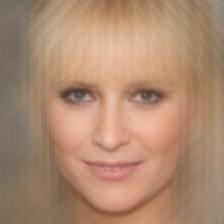} 
    & \includegraphics[width=\linewidth]{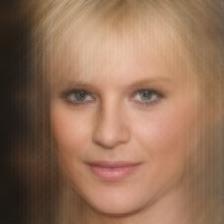}  
    & \includegraphics[width=\linewidth]{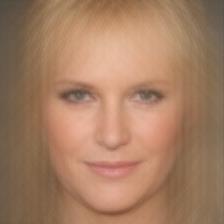}
    & \includegraphics[width=\linewidth]{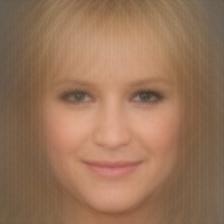}
    & \includegraphics[width=\linewidth]{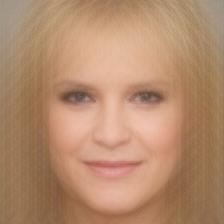}
    & \includegraphics[width=\linewidth]{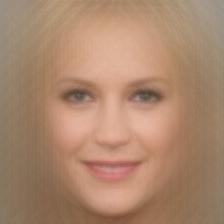}
    & \includegraphics[width=\linewidth]{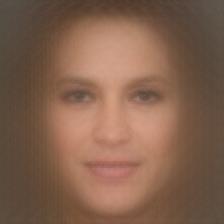}
\end{tabular}%
\vspace{-2mm}
\caption{Visualization of input and reconstructed images with different $\lambda_{2}$ (shown on top of each column of images). Other hyperparameters are fixed: $\lambda_1\,{=}\,1, \mu_1\,{=}\,0,\mu_2\,{=}\,1$. 
}
\label{fig:perceptual_comparison}
\vspace{-3mm}
\end{figure*}

\para{More Data.}
We validate the proposed model by adding more data (10k images from MS-Celeb-1M) to $\mathcal{X}_{2}$ (\#5 and \#6) to train $\DecAdv$. 
While extra data similar to the data distribution of $\mathcal{X}_{1}$ helps the adversary to improve $\DecAdv$, as shown from the face similarity comparisons between \#1 and \#5 (0.851 to 0.865), the $\Enc$ trained with our method is still more robust to the {\INV} attack.
Furthermore, providing the same amount of data for $\Enc$ training (\#7 and \#8) improves the MCC slightly,
while both face similarity and feature similarity decrease more.
This demonstrates that more data for the $\Enc$ training will help protect the privacy. 
The reason is more data would help better simulate a strong fake adversary during the training stage, such that the $\Enc$ can learn to encode images into more secure features.

\para{Single Attribute.}
We consider a single attribute prediction. Compared to 40 attributes prediction, the network needs to maintain much less amount of information for a single attribute prediction, and thus it should be easier to make features privacy-protected. 
%
For example, the baseline $\Enc$ trained for a single attribute prediction (\#9) has significantly lower face and feature similarities compared to that trained for $40$ attributes prediction (\#1). Our method (\#10) further improves the privacy-protection thanks to the proposed negative reconstruction loss. 
%
%
As shown in the rightmost column of Figure~\ref{fig:adversarial_comparison}, the reconstructed images look like the mean faces reflecting only the smiling attribute, as the information of all other attributes is automatically discarded.

\begin{table}[t]
\def\arraystretch{1}
\setlength\tabcolsep{4.5pt}
  \centering
  \caption{Results with different $\lambda_{2}$ in the training stage. Other hyperparameters are fixed: $\lambda_{1}\,{=}\,1, \mu_1\,{=}\,0,\mu_2\,{=}\,1$.}
  \footnotesize
    \begin{tabular}{c|c|c|c|c|c}
    \toprule
    $\lambda_{2}$ & MCC $\uparrow$ &Face Sim. $\downarrow$ & Feat. Sim. $\downarrow$ & SSIM  & PSNR \\
    \hline
    0 & 0.631 & 0.631  & 0.862 & 0.300 & 15.445 \\ 
    \hline
    1 & 0.582 & 0.575  & 0.710  & 0.299 & 15.371 \\ 
    \hline
    2 & 0.455 & 0.545  & 0.604 & 0.273 & 14.920 \\ 
    \hline
    3 & 0.417 & 0.528  & 0.568 & 0.248 & 14.454 \\ 
    \hline
    4 & 0.311 & 0.507  & 0.542 & 0.225  & 14.048 \\ 
    \hline
    5 & 0.255 & 0.502 & 0.530 &  0.224 & 14.047  \\ 
    \hline
    10 & 0.000 & 0.294   & 0.374 & 0.158 &  12.899  \\   
    \bottomrule
    \end{tabular}
    \vspace{-3mm}
  \label{tab:perceptual}
\end{table}

\subsection{Visualization of Latent Representations}

We visualize the learned features using t-SNE~\cite{maaten2008visualizing} to analyze the effectiveness of the proposed method. 
Figure~\ref{fig:latent_visualization} shows the t-SNE plots for two scenarios, either trained for 40 attribute prediction tasks or a single ({\it Smiling}) attribute prediction task. 
For presentation clarity, we colorcode data points with respect to two attributes, namely, {\it Eyeglasses} and {\it Smiling}. 

The baseline model features are well separable along both attributes when trained for 40 attributes. 
On the other hand, the features of the model trained with adversarial reconstruction loss are more uniformly distributed along these two attributes since the encoder is forced to discard the private information relevant to the reconstruction as much as possible. 

For the single attribute case, only the {\it Smiling} attribute is preserved for both models. 
The red and blue dots are mixed in the right side of Figure~\ref{fig:latent_visualization}, which indicates the attribute {\it Eyeglasses} is recognizable or separable. 
That is because the attribute {\it Eyeglasses} is not the target training utility. 
The adversarial reconstruction loss does not turn out to be particularly effective since the retained information in the feature space is already minimal but sufficient for a single attribute prediction and adding an extra loss to interfere the reconstruction does not have a large benefit. 
This also corresponds to the last columns of Figure~\ref{fig:adversarial_comparison}, where the reconstructed images are almost mean faces with only smiling attributes preserved.

%
%

%
%
%

\subsection{Ablation Study of Perceptual Distance Loss}

We analyze the influence of the perceptual distance loss on the privacy preserving mechanism.
To pinpoint its impact, we perform experiments by changing $\lambda_{2}$, which corresponds to the weight of the perceptual distance term in~\eqref{eq:negative_perceptual_loss}, while fixing other hyperparameters for $\Enc$ ($\lambda_1\,{=}\,1$) and $\DecAdv$ ($\mu_1\,{=}\,0, \mu_2\,{=}\,1$).
All utility and privacy metrics are reported in Table~\ref{tab:perceptual} and the reconstructed images are presented in Figure~\ref{fig:perceptual_comparison}.


Changing $\lambda_{2}$ clearly demonstrates the trade-off between utility and privacy. As we increase $\lambda_{2}$, the model becomes more privacy-protecting as demonstrated by the lower face and feature similarities, but this comes at the cost of decreased utility performance (mean MCC). 
%
Our quantitative analysis is also consistent with the visual results, where the reconstructed images in Figure~\ref{fig:perceptual_comparison} contain less sensitive information as $\lambda_{2}$ increases and tend to be a mean face when $\lambda_{2}$ becomes as high as $10$. 
The trade-off between utility and privacy suggests that we can achieve different level of privacy in real applications.


\begin{table*}[t]
\def\arraystretch{1}
\setlength\tabcolsep{3.5pt}
 \centering
 \caption{Evaluation on different layers. We show that our method ``w/ adv'' maintains a good mean MCC while reducing the face similarity to protect the privacy, consistently improving across all feature layers compared to ``w/o adv''.
Furthermore, our method shows smaller LDA scores, which indicate that the relative distance among the features of different identities becomes smaller, hence benefiting privacy preservation.
}
 \scalebox{0.9}{
    \begin{tabular}{lrr|rr|rr|rr|rr|rr}
    \toprule
    \multirow{2}[0]{*}{} & \multicolumn{2}{c}{Conv1} & \multicolumn{2}{c}{Conv2} & \multicolumn{2}{c}{Conv3} & \multicolumn{2}{c}{Conv4} & \multicolumn{2}{c}{Avg Pool} & \multicolumn{2}{c}{FC Layer} \\
         & w/o adv & w/ adv & w/o adv & w/ adv & w/o adv & w/ adv & w/o adv & w/ adv & w/o adv & w/ adv & w/o adv & w/ adv \\
    \midrule
    Face Sim. $\downarrow$ & 0.98  & 0.36  & 0.95  & 0.48  & 0.74  & 0.49  & 0.64  & 0.52  & 0.55  & 0.50  & 0.54  & 0.51  \\
    Mean MCC $\uparrow$ & 0.64  & 0.65  & 0.65  & 0.64  & 0.64  & 0.64  & 0.65  & 0.64  & 0.65  & 0.61  & 0.64  & 0.61  \\
    Within Var ($S_w$) & 1468.77  & 768.29  & 1543.09  & 1523.31  & 2067.54  & 1779.75  & 2212.66  & 2155.19  & 2209.97  & 2243.49  & 2541.02  & 1061.99  \\
    Between Var ($S_b$) & 2884.42  & 218.88  & 2726.15  & 597.48  & 1608.97  & 598.39  & 1199.69  & 747.37  & 1155.09  & 935.16  & 983.14  & 710.93  \\
    LDA Score $\downarrow$ & 1.96  & 0.28  & 1.77  & 0.39  & 0.78  & 0.34  & 0.54  & 0.35  & 0.52  & 0.42  & 2.58  & 1.49  \\
    \bottomrule
    \end{tabular}%
  }
  \vspace{-2mm}
 \label{tab:different_layers}%
\end{table*}%

\subsection{Ablation Study of Different Layers}
As in~\cite{dosovitskiy2016inverting,dosovitskiy2016generating}, the {\INV} attack becomes much easier when low- or mid-level features are used for inversion.
We analyze the effect of features from different layers on the utility and privacy. 
We choose six different layers of intermediate features in $\Enc$, and adapt the network structures of $\Dec$ and $f$ accordingly. 
Specifically, $\Enc$ and $\Dec$ are symmetric to each other, and $f\circ\Enc$ is the entire ResNet-50 architecture followed by 2 fully connected layers at all cases.

We first compare the face similarity and MCC in Table~\ref{tab:different_layers}.
The face similarity decreases as the layer goes deeper for the baseline model because the information becomes abstract during the forwarding process. 
With adversarial training, the face similarity are reduced compared to those without adversarial training, while the MCC is not affected significantly. 
Furthermore, the face similarity with adversarial training in lower layers is generally lower than that in deeper layers. 

Next, we present within-class variance $S_w$, between-class variance $S_b$ and the LDA score $S_b/S_w$.
A low LDA score indicates that the relative distance among the features of different identities is small, thus more privacy-preserving, whereas the distance between two features of the same identity becomes large. 
As shown in Table~\ref{tab:different_layers}, the LDA score of baseline model decreases as the layer goes deeper. 
In addition, the LDA score of our model is generally smaller than that of the baseline, which further validates that the features with adversarial training are more uniformly distributed.
%

\section{Relation to Information Bottleneck}
\label{sec:theorectical_analysis}
While our method is developed by integrating potential attacks from adversary (e.g., decoding latent representation into the pixel space), our method can be understood from the  information-theoretic perspective.
%
%
The objective function can be mathematically formalized using mutual information and conditional entropy:
\begin{align}
\label{eq:mi_formulation}
\min_{\Enc,f}\max_{\Dec}I(\Dec(Z); X) + H(Y|f(Z)),
\end{align}
where $Z\,{=}\,\Enc(X)$. Note that our objective resembles that of information bottleneck methods~\cite{tishby2000information,achille2018information,alemi2016deep} except that we introduce the decoder to estimate mutual information via a min-max game between $\Enc$ and $\Dec$~\cite{belghazi2018mine}. The mutual information term can be reduced as:
\begin{align}
I(\Dec(Z); X){=}H(X){-}H(X|\Dec(Z)).
\end{align}
Since $H(X)$ is a constant, the protector's objective is:
\begin{align}\label{eq:min-max}
\min_{\Enc,f}\max_{\Dec}-H(X|\Dec(Z)) + H(Y|f(Z)).
\end{align}
If we use the reconstruction error for the first term:
\begin{align}
	H(X|\Dec(Z))=\mathbb{E}_{\{(X,Z)\}}\big[\Vert\Dec(Z)\,{-}\,X\Vert^2\big],
\end{align}
\eqref{eq:min-max} is realized as a minimization of negative reconstruction loss for updating $\Enc$.
If we use reconstruction error and perceptual error for the first term:
\begin{align}
	H(X|\Dec(Z))&= \lambda_1 \mathbb{E}_{\{(X,Z)\}}\big[\Vert\Dec(Z)\,{-}\,X\Vert^2 \nonumber \\
	&+ \lambda_2 \Vert g(\Dec(Z))\,{-}\,g(X)\Vert^2 \big],
\end{align}
\eqref{eq:min-max} is realized as a minimization of negative reconstruction loss and negative perceptual loss for updating $\Enc$ as in \eqref{eq:protector_formula}.

Aside from empirical results, the information-theoretic perspective of our method also reveals some limitations of the current approach. First, maximizing $H(X|\Dec(Z))$ may not directly provide privacy-preservation to latent representations, which is evident from the data-processing inequality $H(X|\Dec(Z))\,{\geq}\,H(X|Z)$.
To protect privacy against universal attacks, it is necessary to develop methods to maximize $H(X|Z)$. 
Furthermore, our algorithm can defend against data reconstruction adversary, but there might be useful private information remaining that adversaries can take advantage of, such as the face ethnicity information from the skin color of reconstructed images, even though the reconstructed images are not recognized the same as the input images.
%
However, the proposed method provides an effective method to protecting against private attributes, which is of great interest towards learning explainable representations in the future. 

\section{Conclusions}
We propose an adversarial learning framework to learn a latent representation that preserves visual data privacy.
Our method is developed by simulating the adversary's expected behavior for the model inversion attack and is realized by alternating update of $\Enc$ and $\Dec$ networks.
We introduce quantitative evaluation methods and provide comprehensive analysis of our adversarial learning method.
Experimental results demonstrate that our algorithm can learn privacy-preserving and task-oriented representations.

\para{Acknowledgements.} This work is supported in part by the NSF CAREER Grant \#1149783 and gifts from NEC.

\bibliographystyle{aaai}
\bibliography{AAAI-XiaoT.1790}

\clearpage

\section*{Appendix}

\begin{figure}[b]
    \centering
    \includegraphics[width=.9\linewidth]{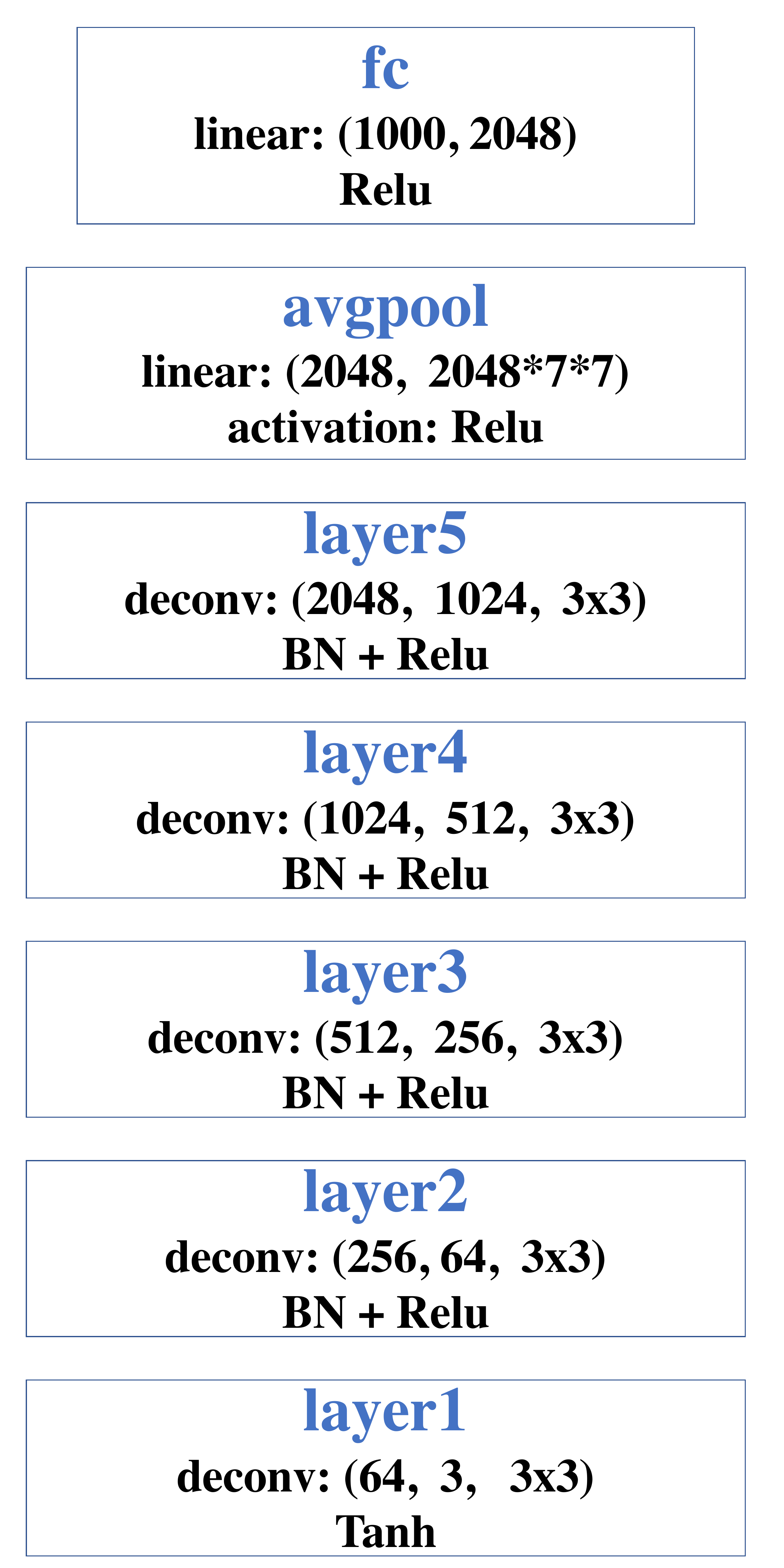}
    \caption{$\Dec$ and $\DecAdv$ network architectures.}
	\label{fig:decoder}
\end{figure}

\section{Network Architectures}

The encoder is set to be the standard ResNet-50 architecture. We refer an official implementation for a detailed reference\footnote{ \url{https://github.com/pytorch/vision/blob/master/torchvision/models/resnet.py}}. Unless otherwise stated, we extract the feature for $\Enc$ from the \texttt{fc} layer. The architecture of the decoder is almost reverse to the $\Enc$ with up-sampling, as shown in Figure~\ref{fig:decoder}.

In the ablation study with different layers of features, we choose six different layers of ResNet-50. The architectures of $\Dec$ and latent classifier $f$ change accordingly. For instance, if we choose \texttt{layer3} as the intermediate layer, then the decoder should start from \texttt{layer3} to \texttt{layer1} and the latent classifier should start from \texttt{layer4} to \texttt{fc} with two additional fully connected layers followed.

\section{Visualization with Perceptual Distance Loss}

\begin{figure*}[!t]
\centering
\def\arraystretch{1}
\begin{tabular}{*{8}{m{0.1\linewidth}<{\centering}}}
    input & 0 & 1 & 2 & 3 & 4 & 5 & 10 \\
      \includegraphics[width=\linewidth]{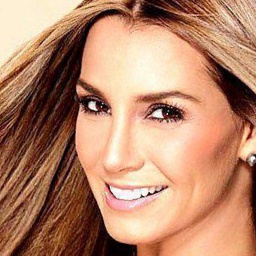}
    & \includegraphics[width=\linewidth]{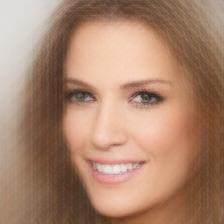} 
    & \includegraphics[width=\linewidth]{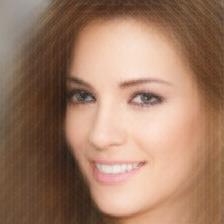}  
    & \includegraphics[width=\linewidth]{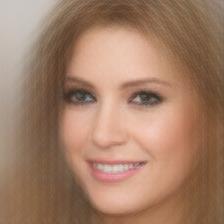}
    & \includegraphics[width=\linewidth]{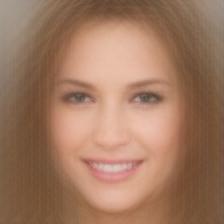}
    & \includegraphics[width=\linewidth]{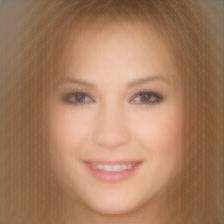}
    & \includegraphics[width=\linewidth]{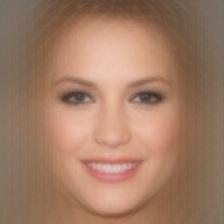}
    & \includegraphics[width=\linewidth]{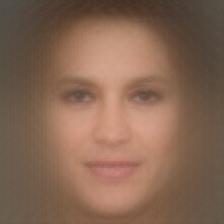} 
    \\
      \includegraphics[width=\linewidth]{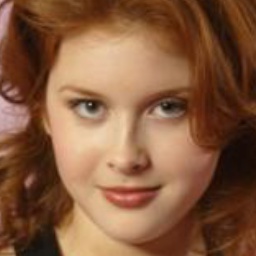}
    & \includegraphics[width=\linewidth]{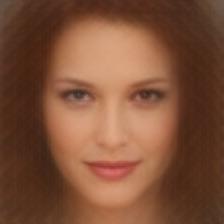} 
    & \includegraphics[width=\linewidth]{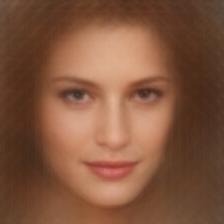}  
    & \includegraphics[width=\linewidth]{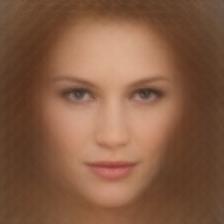}
    & \includegraphics[width=\linewidth]{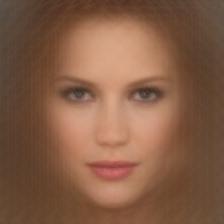}
    & \includegraphics[width=\linewidth]{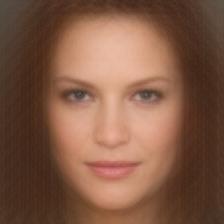}
    & \includegraphics[width=\linewidth]{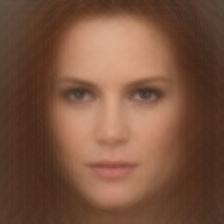}
    & \includegraphics[width=\linewidth]{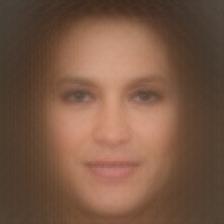}
    \\
      \includegraphics[width=\linewidth]{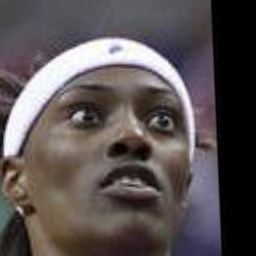}
    & \includegraphics[width=\linewidth]{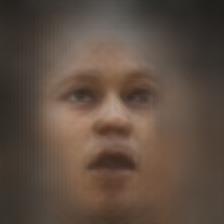} 
    & \includegraphics[width=\linewidth]{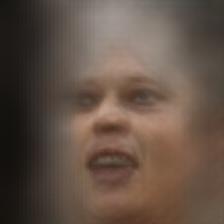}  
    & \includegraphics[width=\linewidth]{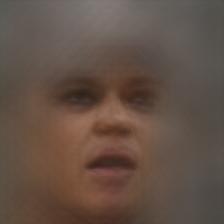}
    & \includegraphics[width=\linewidth]{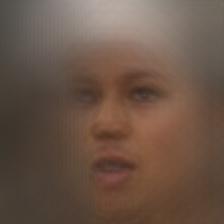}
    & \includegraphics[width=\linewidth]{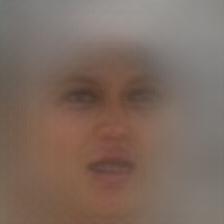}
    & \includegraphics[width=\linewidth]{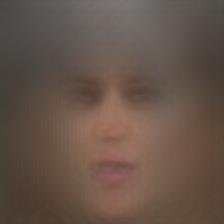}
    & \includegraphics[width=\linewidth]{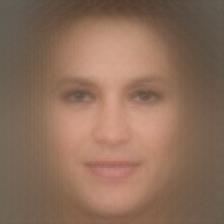}
    \\
      \includegraphics[width=\linewidth]{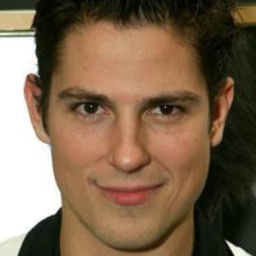}
    & \includegraphics[width=\linewidth]{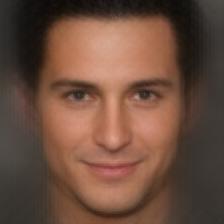} 
    & \includegraphics[width=\linewidth]{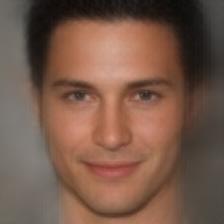}  
    & \includegraphics[width=\linewidth]{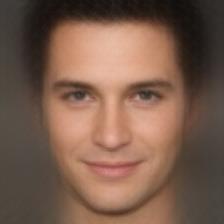}
    & \includegraphics[width=\linewidth]{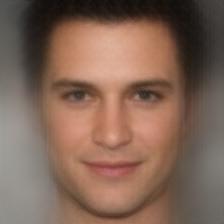}
    & \includegraphics[width=\linewidth]{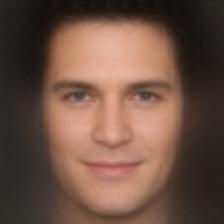}
    & \includegraphics[width=\linewidth]{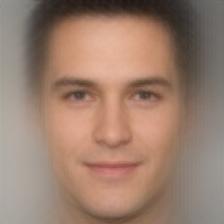}
    & \includegraphics[width=\linewidth]{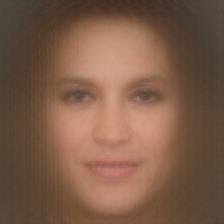}
    \\
      \includegraphics[width=\linewidth]{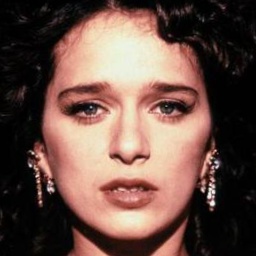}
    & \includegraphics[width=\linewidth]{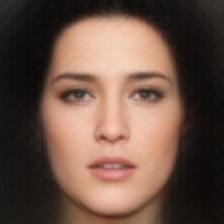} 
    & \includegraphics[width=\linewidth]{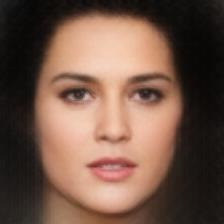}  
    & \includegraphics[width=\linewidth]{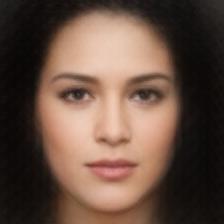}
    & \includegraphics[width=\linewidth]{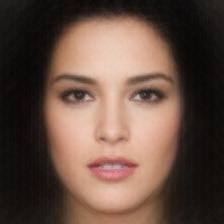}
    & \includegraphics[width=\linewidth]{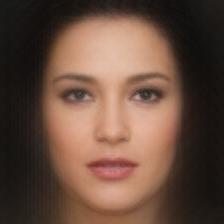}
    & \includegraphics[width=\linewidth]{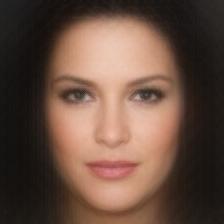}
    & \includegraphics[width=\linewidth]{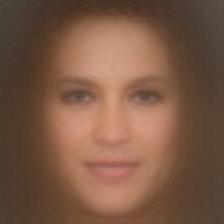}
    \\
      \includegraphics[width=\linewidth]{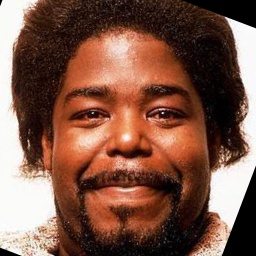}
    & \includegraphics[width=\linewidth]{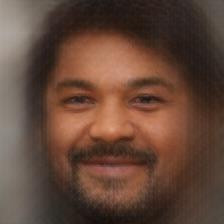} 
    & \includegraphics[width=\linewidth]{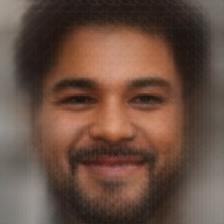}  
    & \includegraphics[width=\linewidth]{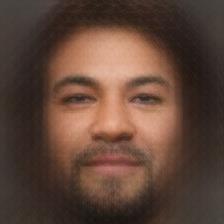}
    & \includegraphics[width=\linewidth]{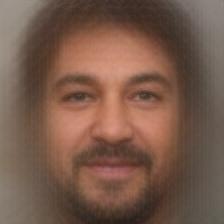}
    & \includegraphics[width=\linewidth]{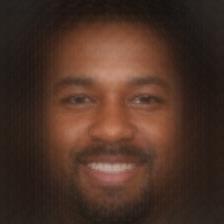}
    & \includegraphics[width=\linewidth]{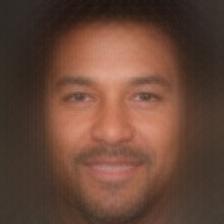}
    & \includegraphics[width=\linewidth]{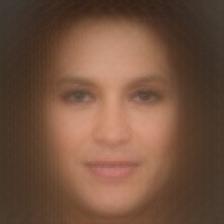}
    \\
      \includegraphics[width=\linewidth]{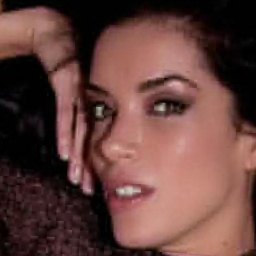}
    & \includegraphics[width=\linewidth]{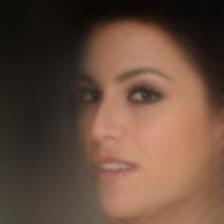} 
    & \includegraphics[width=\linewidth]{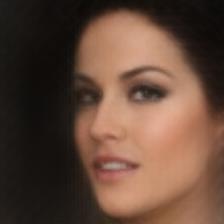}  
    & \includegraphics[width=\linewidth]{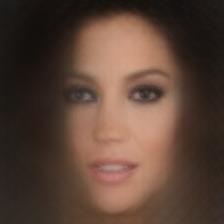}
    & \includegraphics[width=\linewidth]{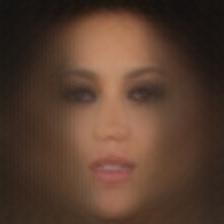}
    & \includegraphics[width=\linewidth]{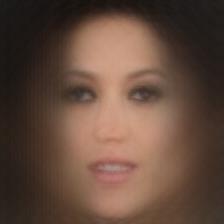}
    & \includegraphics[width=\linewidth]{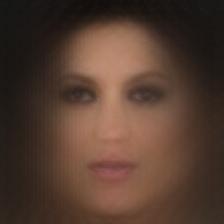}
    & \includegraphics[width=\linewidth]{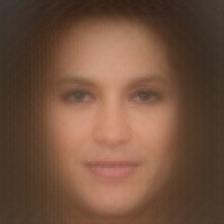}
    \\
      \includegraphics[width=\linewidth]{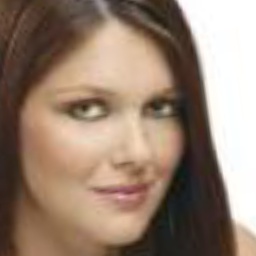}
    & \includegraphics[width=\linewidth]{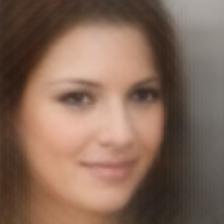} 
    & \includegraphics[width=\linewidth]{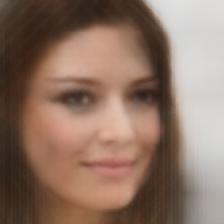}  
    & \includegraphics[width=\linewidth]{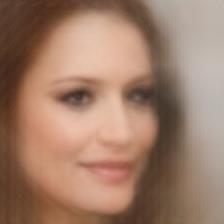}
    & \includegraphics[width=\linewidth]{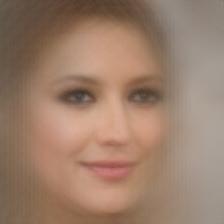}
    & \includegraphics[width=\linewidth]{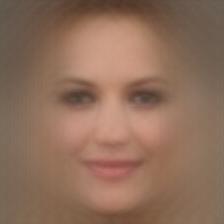}
    & \includegraphics[width=\linewidth]{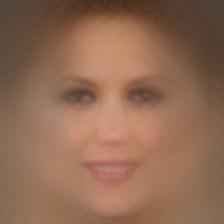}
    & \includegraphics[width=\linewidth]{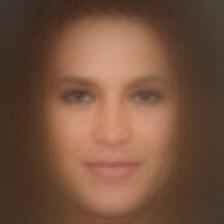}
    \\
      \includegraphics[width=\linewidth]{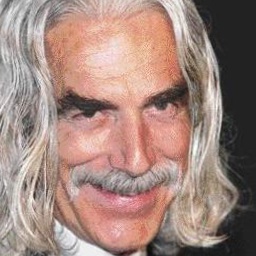}
    & \includegraphics[width=\linewidth]{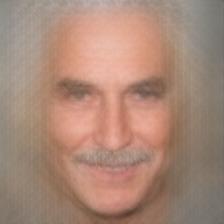} 
    & \includegraphics[width=\linewidth]{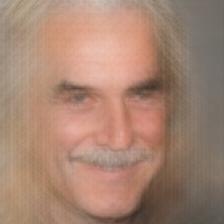}  
    & \includegraphics[width=\linewidth]{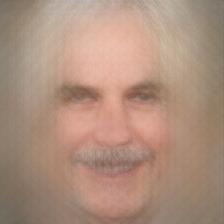}
    & \includegraphics[width=\linewidth]{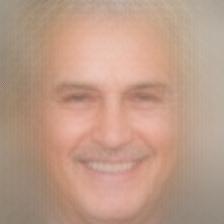}
    & \includegraphics[width=\linewidth]{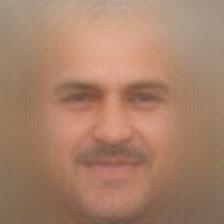}
    & \includegraphics[width=\linewidth]{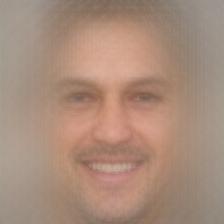}
    & \includegraphics[width=\linewidth]{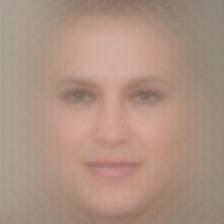}
\end{tabular}%
\caption{Reconstructed images with different $\lambda_{2}$ (shown on top of each image) by fixing $\lambda_1\,{=}\,1$, and $\mu_1\,{=}\,0, \mu_2\,{=}\,1$.
}
\label{fig:perceptual_comparison_supp}
\end{figure*}

In Figure~\ref{fig:perceptual_comparison_supp}, we visualize additional results for reconstruction to demonstrate the trade-off between utility and privacy by changing the regularization coefficient of perceptual distance loss $\lambda_2$ while fixing other hyperparameters for $\Enc$ ($\lambda_1=1)$ and $\DecAdv$ ($\mu_1=0, \mu_2=1$).

\end{document}